\documentclass[journal]{IEEEtran}
\usepackage{algorithm}
\usepackage[english]{babel}
\usepackage{blindtext}
\usepackage{algorithmic}
\usepackage{color, soul}

\usepackage{multirow}
\usepackage{xcolor}
\usepackage{multirow}
\usepackage{xcolor}
\usepackage{mathrsfs}
\usepackage{amsmath,graphicx,setspace,algorithm,algorithmic,multirow,cases,amssymb,subfigure}
\usepackage{amsfonts}
\usepackage{overpic}
\usepackage{epsfig}
\usepackage{graphicx}
\usepackage{amsmath}
\usepackage{amssymb,subfigure,cases}
\usepackage{color,hyperref}
\usepackage{cite}
\usepackage{caption}
\usepackage{multirow}
\usepackage{booktabs}
\usepackage[table]{xcolor}

\usepackage{algorithmic}
\usepackage{stfloats}
\usepackage{colortbl}

\usepackage{lipsum}
\newcounter{sent}
\newcommand{\sent}{\refstepcounter{sent}(\thesent)~}

\ifCLASSINFOpdf
\else
\fi
\definecolor{darkblue}{rgb}{0.0,0.0,1.0}
\hypersetup{colorlinks,breaklinks,
            linkcolor=darkblue,urlcolor=darkblue,
            anchorcolor=darkblue,citecolor=darkblue}

\begin{document}

\title{MSAM: Multi-Semantic Adaptive Mining for Cross-Modal Drone Video-Text Retrieval}

\author{ Jinghao Huang,~\IEEEmembership{}
         Yaxiong Chen,~\IEEEmembership{Member,~IEEE}
         and Ganchao Liu~\IEEEmembership{}

\thanks{Manuscript received XX XX, XXXX.(Corresponding author: Yaxiong Chen.)}

\thanks{
Jinghao Huang is with the School of Computer Science and Engineering Sun Yat-sen University, Guangzhou 510006, China.

Yaxiong Chen is with the School of Computer Science and Artificial Intelligence, Wuhan University of Technology, Wuhan 430070, China, and also with the Sanya Science and Education Innovation Park of Wuhan University of Technology, Sanya 572000.

Ganchao Liu is with the School of Artificial Intelligence, Optics and Electronics (iOPEN), Northwestern Polytechnical University, Xi’an 710072, China.

}}

\maketitle

\begin{abstract}
With the advancement of drone technology, the volume of video data increases rapidly, creating an urgent need for efficient semantic retrieval. We are the first to systematically propose and study the drone video-text retrieval (DVTR) task. Drone videos feature overhead perspectives, strong structural homogeneity, and diverse semantic expressions of target combinations, which challenge existing cross-modal methods designed for ground-level views in effectively modeling their characteristics. Therefore, dedicated retrieval mechanisms tailored for drone scenarios are necessary. To address this issue, we propose a novel approach called Multi-Semantic Adaptive Mining (MSAM). MSAM introduces a multi-semantic adaptive learning mechanism, which incorporates dynamic changes between frames and extracts rich semantic information from specific scene regions, thereby enhancing the deep understanding and reasoning of drone video content. This method relies on fine-grained interactions between words and drone video frames, integrating an adaptive semantic construction module, a distribution-driven semantic learning term and a diversity semantic term to deepen the interaction between text and drone video modalities and improve the robustness of feature representation. To reduce the interference of complex backgrounds in drone videos, we introduce a cross-modal interactive feature fusion pooling mechanism that focuses on feature extraction and matching in target regions, minimizing noise effects. Extensive experiments on two self-constructed drone video-text datasets show that MSAM outperforms other existing methods in the drone video-text retrieval task. The source code and dataset will be made publicly available.
 \end{abstract}


\IEEEpeerreviewmaketitle

\section{Introduction}

\begin{figure}
\begin{center}
\includegraphics[width=0.5\textwidth]{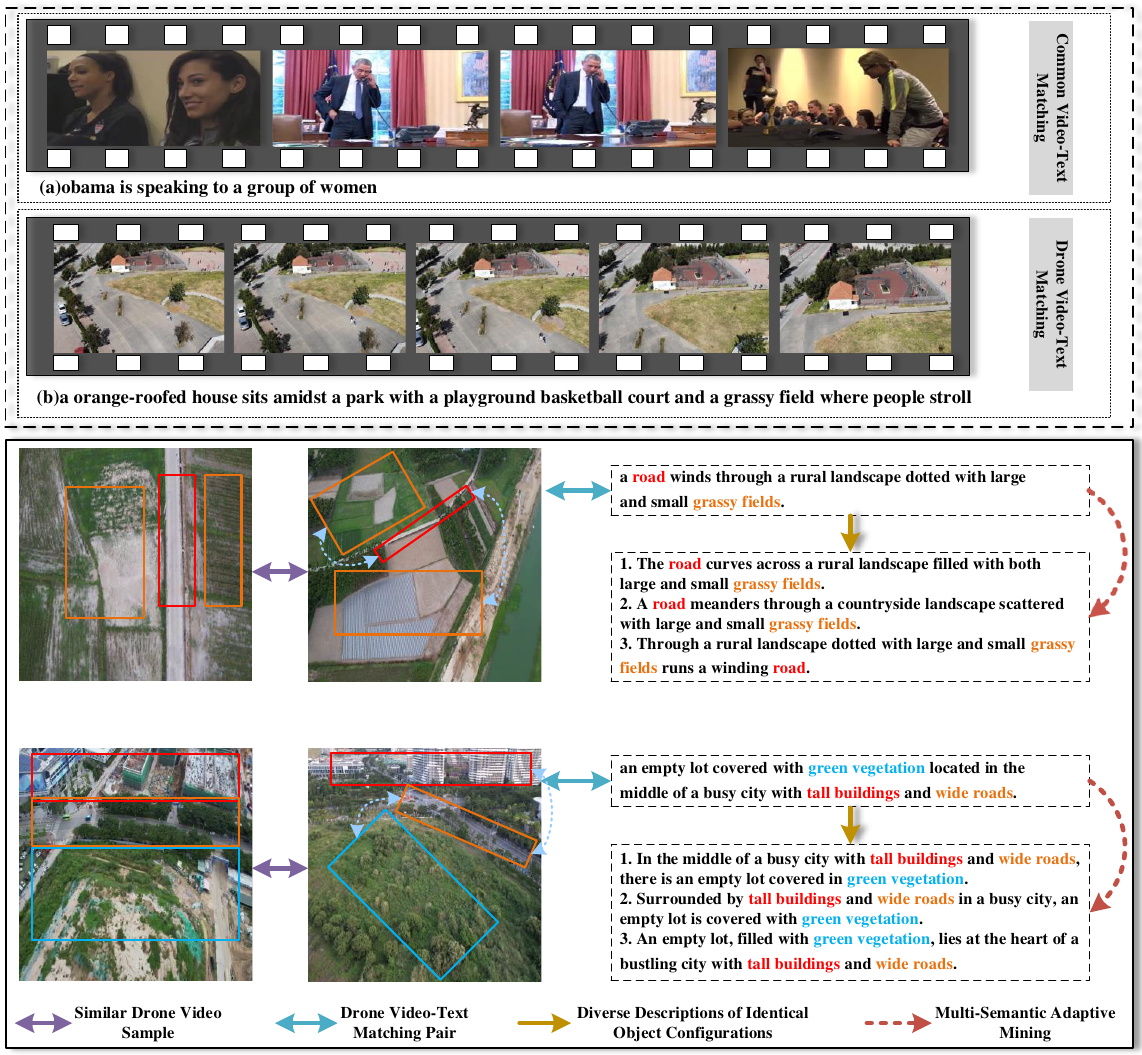}
\renewcommand{\figurename}{Fig.}
\end{center}
\caption{\textbf{Top row:} Comparism of common video-text matching and drone video-text matching. \textbf{Bottom row:} The video frames from the same scene in the drone dataset illustrate the strong structural homogeneity of drone videos, while the target combinations within a single video exhibit semantic diversity in their descriptions.}
\label{fig:n}
\end{figure}

\IEEEPARstart{U}{nmanned} Aerial Vehicles, commonly referred to as drones, often face negative portrayal in media. However, with the rapid advancement of drone technology \cite{2015Overview,2014Unmanned}, drone images have found widespread applications across various domains. These applications include urban planning, agricultural monitoring \cite{2015UAVs,Dong2019Sustainable}, natural disaster monitoring and rescue \cite{2019Evaluation}, as well as urban traffic management \cite{DBLP:conf/cvpr/ZhangZ0LR22,DBLP:journals/tcsv/ChenJZZL24}. In comparison to satellites \cite{DBLP:journals/corr/abs-2210-04936}, drones offer advantages such as lower costs, real-time high-resolution video capabilities, and the ability to make swift decisions through real-time streaming \cite{DBLP:journals/tip/FengLZZJ23}. Furthermore, drones are not limited by weather conditions like cloud cover, granting them a high degree of flexibility. The escalating volume of drone-related data underscores the necessity for effective information extraction, making retrieval technology an essential tool. The goal of retrieval technology is to provide results that best match the query information. However, literature pertaining to drone video analysis remains relatively sparse, primarily focusing on vehicle or human detection \cite{DBLP:journals/tcsv/LengMZGLG23} and tracking \cite{DBLP:journals/tcsv/WuSJK25,DBLP:journals/tcsv/ChenZCWH23}, as well as understanding human activities in relatively controlled environments \cite{DBLP:journals/tcsv/TranBNN24,DBLP:journals/tgrs/HuangCXL24a,DBLP:journals/tgrs/ChenHMJXZ23}. To drive the development of multi-modal drone technologies, this paper introduces a novel task to the drone community, the drone video-text retrieval.

According to the top row of Fig. \ref{fig:n}, drone videos significantly differ from common videos \cite{DBLP:conf/cvpr/LiLZNWL21} in terms of content and shooting perspective. While common videos focus on human behavior and the alignment of temporal sequences, drone videos emphasize understanding scene structure and target objects. Although cross-modal retrieval has achieved considerable progress in common video domains, existing methods encounter significant challenges when directly applied to drone video tasks. Drone videos are typically captured from an overhead perspective, resulting in visual structures that differ substantially from ground-view videos. These videos often contain repetitive elements such as roads, buildings, and vehicles, leading to high visual similarity across different samples, as observed in the bottom row of Fig. \ref{fig:n}. Such visual homogeneity increases the likelihood of multiple videos sharing similar or identical textual descriptions, thereby intensifying cross-modal ambiguity. More critically, due to variations in shooting angles and scene compositions, the same set of target objects may correspond to diverse semantic expressions in different contexts, introducing substantial semantic uncertainty. This makes semantic alignment in DVTR particularly difficult. However, most existing cross-modal methods are designed based on ground-view scenarios and fail to adequately capture the structural characteristics and semantic ambiguity inherent in drone videos, resulting in degraded retrieval performance. Therefore, a cross-modal retrieval method that specifically addresses high similarity and focuses on target region features is needed.


To address these challenges, we can fully utilize the diverse scene features present in drone video frames and explore the multiple combinations between video frames and text descriptions, as observed in the bottom row of Fig. \ref{fig:n}. By extracting the contextual information embedded in the visual cues of the video, we can enhance collaboration between the drone video and text modalities. This improves the capture of semantic relationships between video frames and text, reducing alignment errors caused by intra-class similarity.

\begin{figure*}[htb]
\begin{center}
\centering
\includegraphics[width=0.9\textwidth]{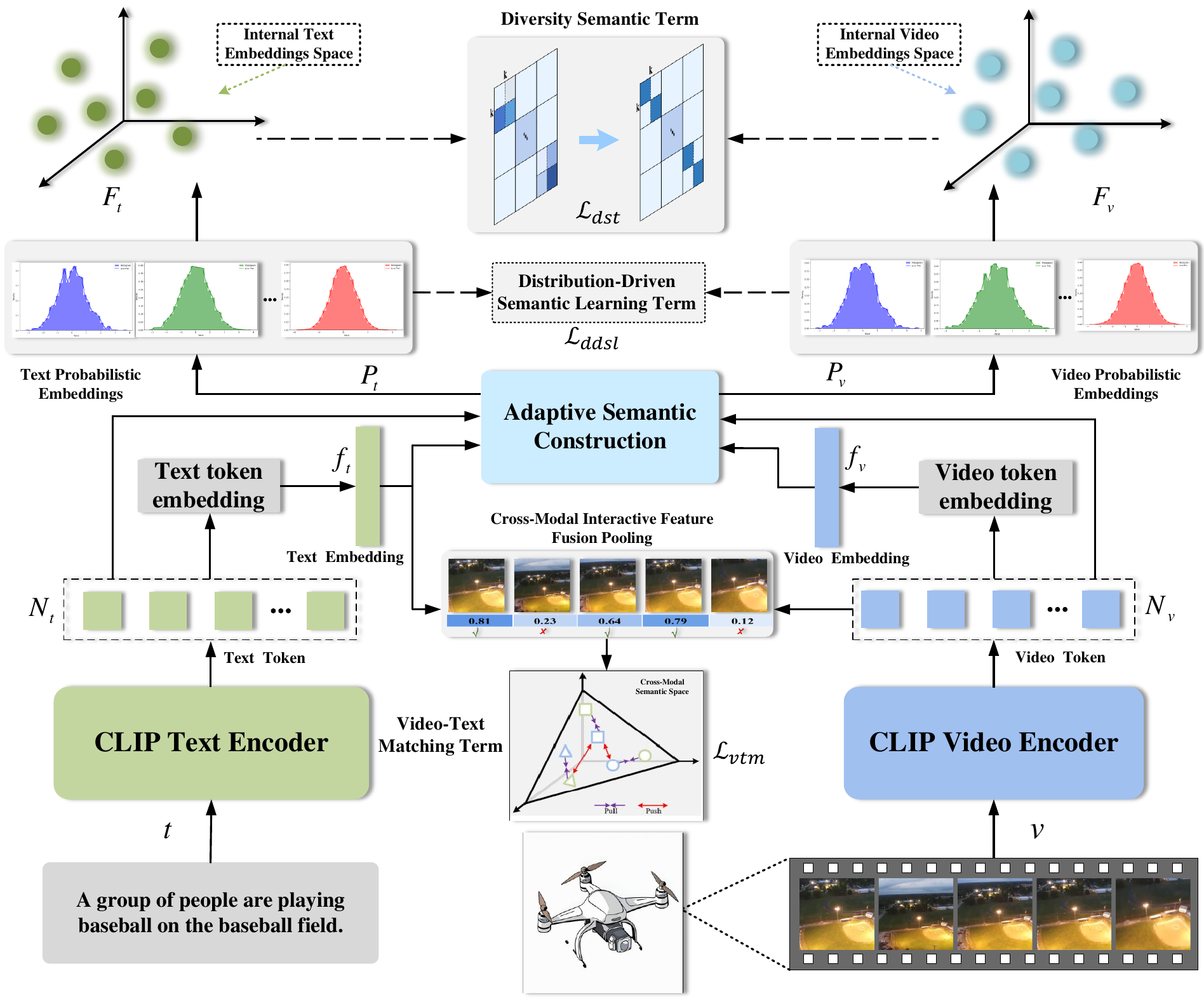}
\renewcommand{\figurename}{Fig.}
\end{center}
\caption{The overall overview of the MSAM framework is as follows. For a given drone video-text pair $P = {\{\boldsymbol{v}, \boldsymbol{t}\}}$, we first encode it using the text encoder $\mathcal C_t$ to obtain $\boldsymbol{N_t}$ and $\boldsymbol{f_t}$. Similarly, the image encoder $\mathcal C_v$ is used to encode the frames of the given video $\boldsymbol{v}$, generating $\boldsymbol{N_v}$ and $\boldsymbol{f_v}$. Then the similarity between $\boldsymbol{N_v}$ and $\boldsymbol{f_t}$ is computed by Cross-Modal Interactive Feature Fusion Pooling and optimize through $\mathcal L_{vtm}$. Additionally, we perform multi-semantic adaptive learning mechanism on all visual frames $\boldsymbol{N_v}$ and text tokens $\boldsymbol{N_t}$ through adaptive semantic construction to generate $k$ probabilistic embeddings. Finally, we use $\mathcal L_{dst}$ and $\mathcal L_{ddsl}$ as constraints to construct multi-instance comparisons within a batch.}
\label{fig:main}
\end{figure*}

We propose an innovative cross-modal approach called Multi-Semantic Adaptive Mining (MSAM), depicted in Fig. \ref{fig:main}. This method extracts semantic features from specific regions, such as roads and buildings. It combines inter-frame variation information to generate rich, multi-layered semantic representations that support deep understanding and reasoning. The core of MSAM consists of an adaptive semantic construction module, a distribution-driven semantic learning term and a diversity semantic term. The adaptive semantic construction module represents each modal sample as a probability distribution, enabling probability alignment instead of traditional deterministic matching. To enhance semantic consistency and facilitate information flow, we design a distribution-driven semantic learning constraint that narrows the distribution gap between drone videos and text descriptions. Additionally, we incorporate a diversity semantic constraint to prevent feature degradation and encourage the learning of diverse semantic representations. To mitigate the interference from complex backgrounds, we propose a cross-modal interactive feature fusion pooling (CIFFP) mechanism. It focuses on extracting features from target-relevant regions while filtering out unrelated background frames. This mechanism supports joint reasoning between text and video frames, optimizing the attention mechanism of the model attention mechanism. Unlike traditional methods, our method reconstructs video embedding features by integrating both text and video information. It effectively filters out frames that are irrelevant to the text description, guiding the model to focus on core semantic information and enhancing the supervisory signal. The primary contributions of this work are as follows:
\begin{enumerate}
    \item We propose a multi-semantic adaptive learning mechanism to adaptively generate multi-semantic embeddings for each drone video and text. This mechanism promotes semantic consistency alignment and enhances cross-modal information flow, thereby improving diversity of the model in semantic learning.
   \item We introduce a cross-modal interactive feature fusion pooling mechanism that enables the model to infer the drone video frames most relevant to the given text. This guides the model to focus on the core semantic content outlined in the text, thereby optimizing the supervision signal.
  \item In this work, we create two unique drone video-text datasets. They are the first of their kind in this field, providing standard evaluation benchmarks for our method and other comparative approaches.
\end{enumerate}

The remainder of the paper is organized as follows: In Section \ref{relatedworks}, we introduce the related work. Then in Section \ref{The Proposed method}, the relevant details of the model is provided. Section \ref{experiments} presentes a large number of experiments to demonstrate the state-of-the-art performance of the model. Finally, in Section \ref{Conclusion}, a brief summary of this paper is provided.


\section{Related Works}\label{relatedworks}
In this section, we focus on summarizing three research directions: video-text retrieval, probabilistic representation, and visual-text representation learning. Video-text retrieval is the main research problem of this paper, while probabilistic representation and visual-text representation learning are the core aspects of our proposed MSAM method.

\subsection{Video-Text Retrieval}

Text-to-video retrieval is a sophisticated task necessitating an in-depth comprehension of videos. Initial investigations leveraged specialized models to derive video features \cite{DBLP:conf/eccv/Gabeur0AS20,DBLP:journals/corr/abs-2307-12545,DBLP:journals/tcsv/ChenDLY24,DBLP:conf/emnlp/XuG0OAMZF21,DBLP:conf/cvpr/WangZ021}. For example, MMT \cite{DBLP:conf/eccv/Gabeur0AS20} utilized a multimodal Transformer to integrate features obtained from video experts. Conversely, T2VLAD \cite{DBLP:conf/cvpr/WangZ021} employed various experts to process raw videos, creating global expert features for overall alignment and assembling local features across diverse modalities for detailed alignment. A separate group of text-to-video retrieval techniques adapted insights from the CLIP model, which was pre-trained using extensive text-image datasets. These approaches harmonized text and video modalities via meticulously crafted mapping techniques \cite{DBLP:journals/corr/abs-2104-08860,DBLP:conf/sigir/ZhaoZWY22,DBLP:conf/cvpr/GortiVMGVGY22,DBLP:conf/mm/MaXSYZJ22}. In the context of end-to-end systems like ClipBERT \cite{DBLP:conf/cvpr/LeiLZGBB021} and Frozen \cite{DBLP:conf/iccv/BainNVZ21}, enhancing training efficiency was paramount. CenterCLIP \cite{DBLP:conf/sigir/ZhaoZWY22} developed a token clustering method that targeted the identification of the most significant tokens, concurrently minimizing memory usage. Align\&Tell \cite{DBLP:journals/tmm/WangZZXY23} incorporated a Transformer decoder to refine local alignment, augmenting broader comparisons. X-Pool \cite{DBLP:conf/cvpr/GortiVMGVGY22} synthesized video frames with text attention weights, whereas X-CLIP \cite{DBLP:conf/mm/MaXSYZJ22} advanced a nuanced contrastive learning approach to foster extensive video-text integration. TS2-Net \cite{DBLP:conf/eccv/LiuXXCJ22} pinpointed the top-k informative tokens per frame, encapsulating key semantic elements for precise frame-level cross-modal correlation, marking a sophisticated design strategy.

\subsection{Probabilistic Representation}

The integration of probability theory into machine learning is well-established \cite{DBLP:books/lib/Murphy12}. In the field of computer vision, HIB \cite{DBLP:conf/iclr/OhMPRSG19} pioneered the use of probabilistic embeddings. These embeddings encapsulate the inherent uncertainty within image representations, addressing the challenges associated with one-to-many mappings in deep metric learning. This innovative strategy has proven effective across a variety of tasks, including face recognition \cite{DBLP:conf/iccv/ShiJ19,DBLP:conf/cvpr/ChangLCW20} and pose estimation \cite{DBLP:conf/eccv/SunZCSA020}. PCME \cite{DBLP:conf/cvpr/ChunORKL21} harnessed probabilistic embeddings to facilitate text-image retrieval, enabling one-to-many associations across a spectrum of visual concepts. Given the dynamic temporal nature of video content, it inherently carries a richer semantic depth. UATVR \cite{DBLP:conf/iccv/FangWLZS0SJW23} conceptualized the retrieval process as an alignment between probabilistic distributions, treating the outputs from diverse modalities as unique probability distributions. Nonetheless, the methods discussed herein merely replicate many-to-many correspondences through video-text pairings and do not adequately account for the intricate constraints governing each set of probabilistic distributions.

\subsection{Visual-Text Representation Learning}
Visual-text learning has become increasingly prominent, capitalizing on the advancements of deep learning in text-visual domains. Various sub-fields within visual-text learning have shown remarkable progress, including text-visual representation learning \cite{DBLP:journals/corr/abs-2303-02483,DBLP:journals/corr/abs-2212-09737,DBLP:journals/corr/abs-2304-00719}, text-visual generation \cite{DBLP:journals/corr/abs-2301-12959,DBLP:journals/corr/abs-2211-09117,DBLP:conf/cvpr/XuNTLD021}, visual-textual alignment \cite{DBLP:conf/cvpr/VinyalsTBE15,DBLP:journals/pami/KarpathyF17,DBLP:conf/iccv/LiZLLF19}, and more. Zhou \emph{et al.} \cite{DBLP:conf/cvpr/ZhouYL022} proposed a conditional prompt learning method for vision-language models. This method guides the learning process of model  by providing conditional prompt information, thereby enhancing its visual-language understanding capability. Ge \emph{et al.} \cite{DBLP:conf/cvpr/GeGLLSQL22} proposed a method that combines video-text retrieval with multiple-choice questions. The introduction of multiple-choice questions guides the video-text retrieval task, leading to improved retrieval accuracy and efficiency. Liang \emph{et al.} \cite{DBLP:conf/cvpr/LiangWZY22} introduced a visual abductive reasoning method. This approach incorporates visual deductive reasoning processes to infer latent information and relationships within images, thereby enhancing image comprehension and reasoning abilities.


\section{The Proposed Method}\label{The Proposed method}
This section details the MSAM model, with its architecture shown in Fig. \ref{fig:main}. The MSAM model is explained in four parts: (\ref{Problem Definition}) Problem Definition, (\ref{Video-Text Encoder}) Video-Text Encoder, (\ref{Cross-Modal Interactive Feature Fusion Pooling}) Cross-Modal Interactive Feature Fusion Pooling and (\ref{Objective Function}) Objective Function.

\subsection{Problem Definition}\label{Problem Definition}
The main objective of the cross-modal DVTR task is to learn a suitable projection function that aligns the representations of drone video and text modalities. It also ensures that similar drone videos and texts exhibit high similarity in the learned feature space.


\subsection{Video-Text Encoder}\label{Video-Text Encoder}
We choose the high-performance, user-friendly CLIP as our base model to enable fairer comparisons with recent studies using the same model \cite{DBLP:journals/tip/YangLZWZGYCJ23}. To achieve CLIP-based text-to-video retrieval, we map text and video frames to a shared latent space and then aggregate the frame embeddings to form a video representation \cite{DBLP:conf/mcpr2/Portillo-Quintero21}. Leveraging the rich semantic associations CLIP has already learned, we build a new latent space to match text with video.

\textbf{Text Encoder} In terms of extracting text representations, we use the text encoding network of CLIP to embed sentences into a $D$-dimensional space. This process facilitates the incorporation of contextual information from words during the encoding phase, thus allowing for a more holistic understanding and processing of the entire sentence. The process is as follows:
\begin{equation}\label{Eq:6}
    \boldsymbol{f_{t}}, \boldsymbol{N_{t}} =\mathcal C_t(\boldsymbol{t};\boldsymbol{\theta_t})
\end{equation}
where $\mathcal C_t(\boldsymbol{x})$ denotes the text encoding network of CLIP. $\boldsymbol{N_t} \in \mathbb{R}^{T \times L \times D}$ represents the visual features of the text extracted by the $\mathcal C_t$, and $\boldsymbol{\theta_t}$ denotes the pre-trained parameters of the $\mathcal C_t$. $T$ represents the number of text samples per batch. $L$ represents the sequence length. $\boldsymbol{f_{t}} \in \mathbb{R}^{T \times D}$ represents the feature vector of the text, where $D$ represents the dimension of the embedded feature space.

\textbf{Video Encoder} We utilize a the visual encoding network of CLIP as the visual feature extractor. This enables us to extract visual features from drone videos, represented as:
\begin{equation}\label{Eq:5}
    \boldsymbol{f_{v}}, \boldsymbol{N_{v}}=\mathcal C_v(\boldsymbol{v}; \boldsymbol{\theta_v})
\end{equation}
where $\mathcal C_v(\boldsymbol{x})$ denotes the visual encoding network of CLIP. $\boldsymbol{f_{v}} \in \mathbb{R}^{B \times D}$ represents the visual features of the drone video extracted by the $\mathcal C_v$, $B$ represents the number of video samples per batch, $\boldsymbol{N_{v}} \in \mathbb{R}^{B \times F \times D}$ represents the feature encoding of $F$ frames in the drone video and $\boldsymbol{\theta_v}$ denotes the pre-trained parameters of the $\mathcal C_v$. 

\subsection{Cross-Modal Interactive Feature Fusion Pooling}\label{Cross-Modal Interactive Feature Fusion Pooling} 
Existing methods often use text-independent temporal aggregation, such as average pooling and self-attention, when processing drone video frames. However, these methods fail to align text with drone video content effectively \cite{DBLP:journals/tip/XuCWDX22}, often encoding information irrelevant to the text description. While X-Pool incorporates cross-attention to fuse text information, it still faces limitations in handling the visual diversity of drone videos. For instance, scenes in drone videos change frequently, with key objects becoming obscured, disappearing, or reappearing as noise \cite{DBLP:journals/tip/YangCZ23}. Therefore, we aim to design a retrieval model that not only integrates text information but also emphasizes inter-frame visual relationships. Specifically, firstly the similarity between each drone video frame embedding $\boldsymbol{N_v}$ and each text embedding $\boldsymbol{f_t}$ is computed. The resulting similarity matrix is processed using the softmax function to determine the attention weights for each drone video frame. The mathematical expression for this operation is:
\begin{equation}\label{Eq:8}
\boldsymbol{S_{v2t}}=softmax(\sum^{D}{\left\|\boldsymbol{f_{t}}\right\|}_F \times {\left\|\boldsymbol{N_{v}}\right\|}_F) 
\end{equation}
where ${\left\|x\right\|}_F$ denotes the Frobenius norm. $\boldsymbol{S_{v2t}} \in \mathbb{R}^{B \times F \times T}$ represents the importance scores of drone video frames after considering the text information, where $T$ denotes the number of text embeddings. The attention weights $\boldsymbol{S_{v2t}}$ are applied to the drone video embeddings $\boldsymbol{N_v}$, enabling the weighted aggregation of the drone video frames. The specific operation is as follows:
\begin{equation}\label{Eq:8}
\boldsymbol{N_{v2v}}=\sum^{F}{\boldsymbol{S_{v2t}} \times \boldsymbol{{\left\|\boldsymbol{N_{v}}\right\|}}_F}
\end{equation}
where $\boldsymbol{N_{v2v}} \in \mathbb{R}^{B \times T \times D}$ represents the aggregated features of each drone video under the influence of the text information. Next, the similarity between the aggregated drone video features $\boldsymbol{N_{v2v}}$ and the text embedding $\boldsymbol{f_t}$ is computed, and the attention weights for the text are obtained using the softmax function, as follows:
\begin{equation}\label{Eq:8}
\boldsymbol{S_{t2v}}=softmax(\sum^{D}{\boldsymbol{N_{v2v}} \times \boldsymbol{{\left\|\boldsymbol{f_{t}}\right\|}}_F})
\end{equation}
where $\boldsymbol{S_{t2v}} \in \mathbb{R}^{B \times T \times 1}$ represents the text relevance scores after incorporating drone video information. Using these text attention weights $\boldsymbol{S_{t2v}}$, the drone video pooled features $\boldsymbol{N_{v2v}}$ are weighted and aggregated, with the specific formula as follows:
\begin{equation}\label{Eq:8}
\boldsymbol{N_{t2v}}=\sum^{T}{\boldsymbol{S_{t2v}} \times \boldsymbol{N_{v2v}}}
\end{equation}
where $\boldsymbol{N_{t2v}} \in \mathbb{R}^{B \times D}$ represents the refined features of each drone video obtained after incorporating drone video information. $\boldsymbol{N_{t2v}}$ is expanded and reshaped to the form $\mathbb{R}^{B \times T \times D}$. Next, the text-weighted drone video features $\boldsymbol{N_{t2v}}$ are further fused with the initially aggregated drone video features $\boldsymbol{N_{v2v}}$ to combine the drone video and text embeddings. Subsequently, the drone video features $\boldsymbol{N_{v2v}}$ are mapped into a single attention score through a fully connected layer. These scores are used to evaluate the importance of each drone video frame and to adjust the fusion ratio of drone video and text embeddings accordingly. Then, the activation function $sigmoid(\boldsymbol{x})$ is applied to normalize these attention scores to the range [0, 1], and they are expanded to the same dimensions as the video embeddings. The specific expression is as follows:
\begin{equation}\label{Eq:8}
\boldsymbol{S_v}=sigmoid(\mathcal F_p(\boldsymbol{N_{v2v}}))
\end{equation}
where $\boldsymbol{S_v} \in \mathbb{R}^{B \times T \times 1}$ represents the fused attention weight. $\mathcal F_p(\boldsymbol{x})$ denotes a fully connected layer. Based on this weight, the drone video and text embeddings are weighted and fused. If a certain drone video frame has a higher attention score, the weight of its drone video embedding is greater in the final result. Conversely, if the attention score is lower, the weight of the drone video embedding is smaller, and the weight of the text embedding is relatively higher. Finally, the fused embeddings are combined with the original drone video embeddings through a residual connection, as expressed by the following formula:
\begin{equation}\label{Eq:14}
   \boldsymbol{N_{vv}}=\boldsymbol{N_{v2v}} + (\boldsymbol{S_v} \times \boldsymbol{N_{v2v}} + (1-\boldsymbol{S_v}) \times \boldsymbol{N_{t2v}})
\end{equation}
Finally, based on the fused $\boldsymbol{N_{vv}} \in \mathbb{R}^{B \times T \times D}$ and the original text embedding $\boldsymbol{f_t}$, the final similarity score $\boldsymbol{S_{vt}} \in \mathbb{R}^{B \times T}$ between the drone video and text pair is computed through matrix multiplication. This score represents the degree of matching between each drone video and text. The formula is as follows:
\begin{equation}\label{Eq:8}
\boldsymbol{S_{vt}}=\sum^{D}{\boldsymbol{N_{vv}} \times \boldsymbol{{\left\|\boldsymbol{f_{t}}\right\|}}_F}
\end{equation}

\subsection{Objective Function}\label{Objective Function}
\subsubsection{\textbf{Video-Text Matching Term}}  
We input $N$ pairs of text and drone video data $\{(\boldsymbol{v_i}, \boldsymbol{t_i})\}_{i=1}^N$ to train the model. To strengthen the alignment between $\boldsymbol{v_i}$ and $\boldsymbol{t_i}$, we propose a video-text matching term $\mathcal L_{vtm}$, which leverages contrastive learning. This objective is designed to increase the lower bound of mutual information between these paired variables. Specifically, the similarity $\boldsymbol{S_{vt}}$ between the $\boldsymbol{N_{v_{i}}}$ of $\boldsymbol{v_i}$ and the $\boldsymbol{f_{t_{i}}}$ of $\boldsymbol{t_i}$ using the CIFFP mechanism is computed. The cross-entropy loss function from \cite{DBLP:journals/corr/abs-2104-08860} is then used, where the matched text-video pairs are considered positive samples, while all other text-video pairs in the batch are treated as negative samples. $\mathcal L_{vtm}$ is as follows:
\begin{equation}\label{Eq:17}
    \mathcal L_{v2t}=-\frac{1}{N}\sum_{i=1}^{N}log\frac{e^{\boldsymbol{S_{vt}}(i,i)/\tau}}{\sum_{j=1}^{N}e^{\boldsymbol{S_{vt}}(i,j)/\tau}}
\end{equation}

\begin{equation}\label{Eq:17}
    \mathcal L_{t2v}=-\frac{1}{N}\sum_{j=1}^{N}log\frac{e^{\boldsymbol{S_{tv}}(j,j)/\tau}}{\sum_{i=1}^{N}e^{\boldsymbol{S_{tv}}(j,i)/\tau}}
\end{equation}

\begin{equation}\label{Eq:17}
    \mathcal L_{vtm}=\mathcal L_{v2t}+\mathcal L_{t2v}
\end{equation}
where $\tau$ represents a temperature parameter. $\boldsymbol{S_{tv}}$ denotes the transpose of $\boldsymbol{S_{vt}}$.

\begin{figure}[htb]
\begin{center}
\centering
\includegraphics[width=0.4\textwidth]{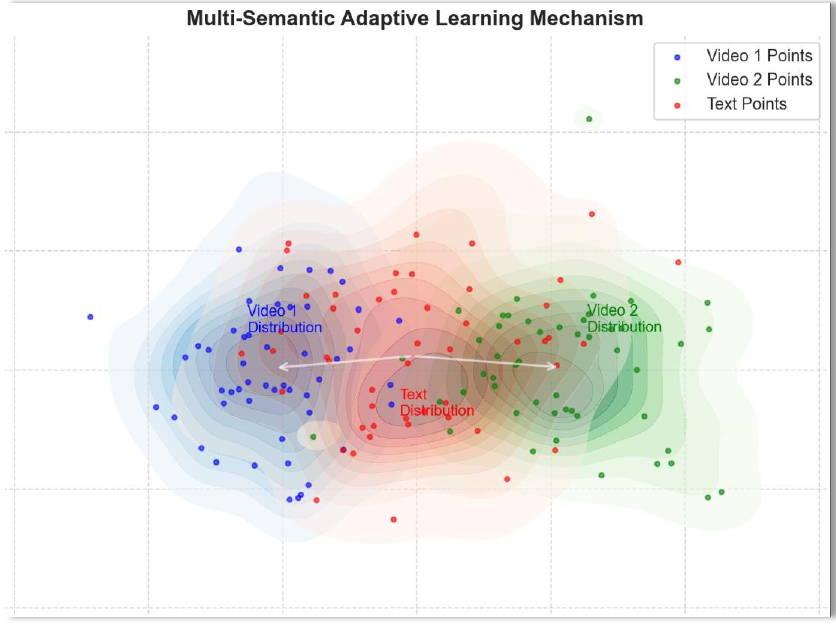}
\renewcommand{\figurename}{Fig.}
\end{center}
\caption{This illustrates MSALM mechanism. The three differently colored shaded areas represent the semantic distributions of two video samples and one text description, reflecting the complex one-to-many relationship between videos and text. The blue and green shades indicate the ambiguous semantic distributions of the videos, while the red shade represents the text distribution. The overlapping areas demonstrate partial matches between the text and multiple videos. The model achieves more flexible and robust semantic matching through multi-point sampling and multi-instance contrastive learning. Light-colored edges emphasize distribution ambiguity.}
\label{fig:msalm}
\end{figure}

\subsubsection{\textbf{Multi-Semantic Adaptive Learning Mechanism}} Drone videos have high similarity within the same category, while deterministic methods typically apply to one-to-one mappings \cite{DBLP:journals/tip/GaoLZSWS22,DBLP:journals/tip/LiuZHZL22}. To address this challenge, we propose the Multi-Semantic Adaptive Learning Mechanism (MSALM), depicted in Fig. \ref{fig:msalm}. It transforms the one-to-one mapping into a one-to-many relationship \cite{DBLP:journals/tip/JinZCZHZ21,DBLP:journals/tip/DingWZHLTLG22}. The method extracts diverse semantics and enhances the model's understanding of varied descriptions and complex scenes. This improves the matching between text and video.

\textbf{Adaptive Semantic Construction} In this method, the attention mechanism is first used to independently optimize the feature outputs $\boldsymbol{N_v}$ and $\boldsymbol{N_t}$. Subsequently, the mapping function $sigmoid(\boldsymbol{x})$ is applied to the distributions of $\boldsymbol{f_v}$ and $\boldsymbol{f_t}$ to enable effective gradient propagation. Next, using LayerNorm \cite{DBLP:journals/corr/BaKH16} and the fully connected layer mapping $\mathcal F_f$, two sets of probabilistic embeddings are generated according to the strategy outlined in reference \cite{DBLP:conf/cvpr/ChunORKL21}. Specifically, $\boldsymbol{H_t} \in \mathbb{R}^{k \times D}$ and $\boldsymbol{H_v} \in \mathbb{R}^{k \times D}$, with each set containing $k$ embeddings. These embeddings satisfy the independent and identically distributed conditions, namely $\boldsymbol{P_t}=\{\boldsymbol{p^{1}_t},\boldsymbol{p^{2}_t},...,\boldsymbol{p^{k}_t}\} \in \mathbb{R}^{k \times D}$ and $\boldsymbol{P_v}=\{\boldsymbol{p^{1}_v},\boldsymbol{p^{2}_v},...,\boldsymbol{p^{k}_v}\} \in \mathbb{R}^{k \times D}$. The specific process is as follows:
\begin{equation}\label{Eq:17}
    \boldsymbol{H_v} = \boldsymbol{f_v} \times sigmoid(\Psi(\boldsymbol{N_v})
\end{equation}
\begin{equation}\label{Eq:17}
    \boldsymbol{H_t} = \boldsymbol{f_t} \times sigmoid(\Psi(\boldsymbol{N_t})
\end{equation}
\begin{equation}\label{Eq:17}
    \boldsymbol{P_v} = \mathcal F_f(\mathcal {LN}(\boldsymbol{H_v}))
\end{equation}
\begin{equation}\label{Eq:17}
    \boldsymbol{P_t} = \mathcal F_f(\mathcal {LN}(\boldsymbol{H_t}))
\end{equation}
where $\Psi(\boldsymbol{x})$ denotes the attention mechanism \cite{c:22}. $\mathcal {LN}(\boldsymbol{x})$ denotes LayerNorm and $\mathcal F_f(\boldsymbol{x})$ denotes a fully connected layer. Subsequently, the vectors $\boldsymbol{P_v}$ and $\boldsymbol{P_t}$ are fed into two deterministic modules, $\mathcal F_{\mu}(\boldsymbol{x})$ and $\mathcal F_{\sigma}(\boldsymbol{x})$. These modules are responsible for predicting the feature mean vectors $\boldsymbol{T_{\mu}}=\{\boldsymbol{\mu^{1}_t},\boldsymbol{\mu^{2}_t},...,\boldsymbol{\mu^{k}_t}\}$ and $\boldsymbol{V_{\mu}}=\{\boldsymbol{\mu^{1}_v},\boldsymbol{\mu^{2}_v},...,\boldsymbol{\mu^{k}_v}\}$, along with the associated data uncertainties $\boldsymbol{T_{\sigma}}=\{\boldsymbol{\sigma^{1}_t},\boldsymbol{\sigma^{2}_t},...,\boldsymbol{\sigma^{k}_t}\}$ and $\boldsymbol{V_{\sigma}}=\{\boldsymbol{\sigma^{1}_v},\boldsymbol{\sigma^{2}_v},...,\boldsymbol{\sigma^{k}_v}\}$. The model uncertainty is represented by the variance of $\{\boldsymbol{\mu^i}\}_{i=1}^{k}$, while data uncertainty is represented by the mean of $\{\boldsymbol{\sigma^{i}}\}_{i=1}^{k}.$ The mean vectors and diagonal covariance matrices are located in the $\mathbb{R}^{D}$ space. The specific computation process is as follows:
\begin{equation}\label{Eq:17}
    \boldsymbol{\mu_t^i},\boldsymbol{\mu_v^i} = \mathcal F_{\mu}(\boldsymbol{p_t^i},\boldsymbol{p_v^i}), i \in [1,k]
\end{equation}
\begin{equation}\label{Eq:17}
    \boldsymbol{\sigma_t^i},\boldsymbol{\sigma_v^i} = \mathcal F_{\sigma}(\boldsymbol{p_t^i},\boldsymbol{p_v^i}), i \in [1,k]
\end{equation}

\textbf{Distribution-Driven Semantic Learning Term} Previous methods \cite{DBLP:conf/iclr/OhMPRSG19,DBLP:conf/cvpr/ChunORKL21} rely on contrastive loss and Monte Carlo estimation to achieve distribution alignment. In contrast, we assume that, given the text and drone video, the latent variable $\boldsymbol{z}$ can be described by two different distributions. Through a unified feature learning or cross-modal learning mechanism, the potential relationships between $\boldsymbol{t}$ and $\boldsymbol{v}$ can be modeled using the latent variable, enabling many-to-many cross-modal matching. The specific process is as follows:
\begin{equation}\label{Eq:17}
    p(\boldsymbol{z}|\boldsymbol{t})=\mathcal N(\boldsymbol{z};\boldsymbol{T_{\mu}},\boldsymbol{T^{2}_{\sigma}}) \sim p(\boldsymbol{z}|\boldsymbol{v})=\mathcal N(\boldsymbol{z};\boldsymbol{V_{\mu}},\boldsymbol{V^{2}_{\sigma}})
\end{equation}
According to this definition, the model updates through a distribution-driven semantic learning term. The training objective is to minimize the difference between the mean and variance of the drone video probability embeddings and those of the corresponding text embeddings. The distribution-driven semantic learning term $\mathcal L_{ddsl}$ is defined as follows:
\begin{equation}\label{Eq:17}
    \mathcal L_{ddsl}=\frac{1}{N}\sum_{i=1}^{N}(log(\frac{\boldsymbol{T_{\sigma}^i}}{\boldsymbol{V_{\sigma}^i}}) - 1 + (\frac{\boldsymbol{V_{\sigma}^i}}{\boldsymbol{T_{\sigma}^i}})^2 + \frac{(\boldsymbol{T_{\mu}^i} - \boldsymbol{V_{\mu}^i})^2}{\boldsymbol{T_{\sigma}^i}^2})
\end{equation}

\begin{figure*}
\begin{center}
\includegraphics[width=0.99\textwidth]{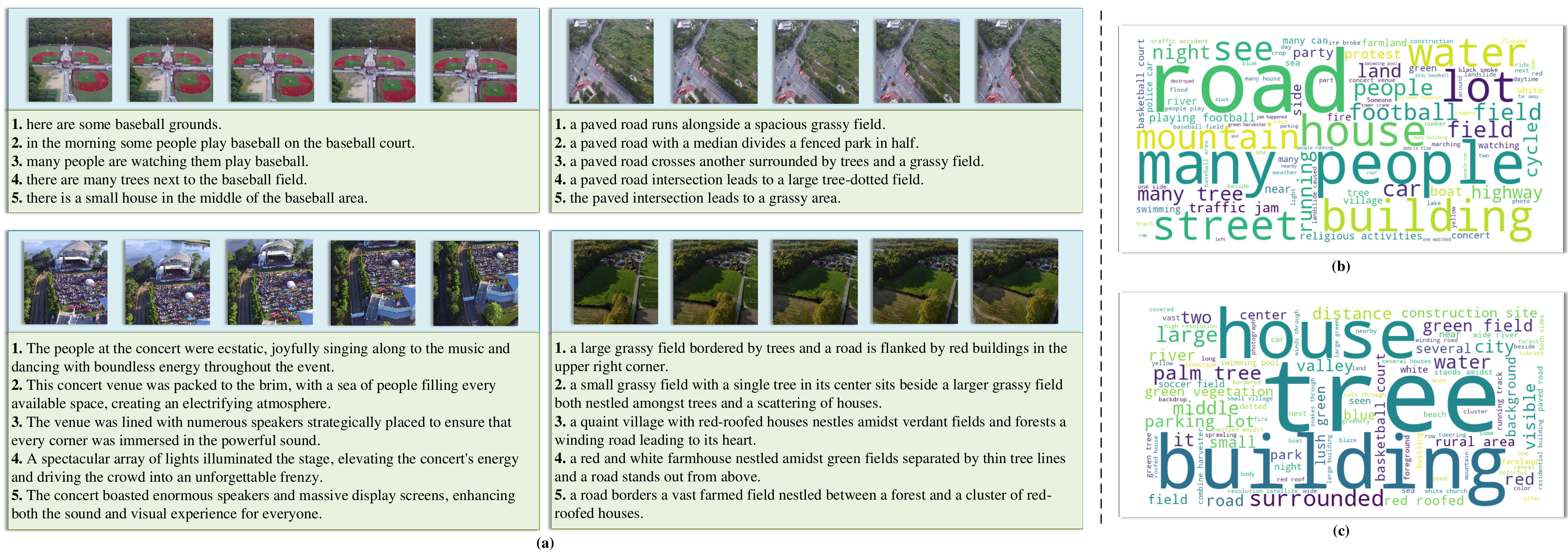}
\renewcommand{\figurename}{Fig.}
\end{center}
\caption{\textbf{Left column:} visualization of the USRD drone video-text dataset and UMCRD drone video-text dataset. \textbf{Right column:} visualization word cloud of the (b) USRD dataset and (c) UMCRD dataset.}
\label{fig:data}
\end{figure*}

\begin{table*}[]
\centering
\caption{Quantitative analysis of USRD and UMCRD drone video-text dataset.}
\scalebox{0.85}{
\begin{tabular}{c|cccccc}
\toprule
\toprule
Dataset  & Video Number    & Video Duration    & Word Number    & Caption Ave Length    & Caption Ave Similarity   & Dataset Diversity    \\ \midrule
USRD  & 2864 & 5s  & 1043 & 10.32  & 0.026 & 3.43  \\
UMCRD & 4235 & 5s$\sim$8s & 2275 & 18.33 & 0.032 & 4.86 \\
\bottomrule
\bottomrule
\end{tabular}
}
\label{table 1}
\end{table*}

\textbf{Diversity Semantic Term}  The objective of consistency semantic alignment is to encourage the model to automatically learn to generate $k$ internal embed features. However, these features may exhibit linear correlation, which limits their capacity to provide rich semantic information. To address this, we apply quality-aware regularization to the embedding features $\boldsymbol{F_t}=\{\boldsymbol{r^{1}_t},\boldsymbol{r^{2}_t},...,\boldsymbol{r^{k}_t}\} = \boldsymbol{T_{\mu}}/\boldsymbol{T_{\sigma}} \in \mathbb{R}^{k \times D}$, $\boldsymbol{F_v}=\{\boldsymbol{r^{1}_v},\boldsymbol{r^{2}_v},...,\boldsymbol{r^{k}_v}\} = \boldsymbol{V_{\mu}}/\boldsymbol{V_{\sigma}} \in \mathbb{R}^{k \times D}$. To ensure that each embedding feature corresponds to different semantic information, it is necessary to enhance the diversity of these features. This objective is achieved by imposing orthogonal constraints, allowing the Gram matrix of these features to approximate the identity matrix under the Frobenius norm. The diversity semantic term for $\mathcal{L}_{dst}$ is defined as follows:
\begin{equation}\label{Eq:8}
    \mathcal L_{dst\_t}=\frac{1}{N}\sum_{i=1}^{N}{\left\|\boldsymbol{F^i_t}{\boldsymbol{F^i_t}}^{T}-\boldsymbol{E}\right\|}_F
\end{equation}
\begin{equation}\label{Eq:8}
    \mathcal L_{dst\_v}=\frac{1}{N}\sum_{i=1}^{N}{\left\|\boldsymbol{F^i_v}{\boldsymbol{F^i_v}}^{T}-\boldsymbol{E}\right\|}_F
\end{equation}
\begin{equation}\label{Eq:17}
    \mathcal L_{dst}=\mathcal L_{dst\_t}+\mathcal L_{dst\_v}
\end{equation}
where $\boldsymbol{E}$ represents an identity matrix. 

\subsubsection{\textbf{Total Objective Term}} Taking into account the considerations from the above three parts, we define $\mathcal L$ as the final objective function, which is formulated as follows:
\begin{equation}\label{Eq:18}
  \mathcal L=\mathcal L_{vtm}+\mathcal L_{ddsl}+{\lambda}\mathcal L_{dst}
\end{equation}
where $\lambda$ is a trade-off parameters.

\begin{table*}[]
\centering
\caption{The experiments on the USRD drone video-text dataset compare MSAM with other state-of-the-art methods.}
\begin{tabular}{c|c|ccccc|ccccc}
\toprule
\toprule
\multirow{2}{*}{Methods} & \multirow{2}{*}{Date} & \multicolumn{5}{c|}{Text $\rightarrow$ Video}                                        & \multicolumn{5}{c}{Video $\rightarrow$ Text}                                       \\
                                & \multicolumn{1}{l|}{}                      & R@1$\uparrow$   & R@5$\uparrow$   & R@10$\uparrow$  & \multicolumn{1}{l}{MdR$\downarrow$} & \multicolumn{1}{l|}{MnR$\downarrow$} & R@1$\uparrow$   & R@5$\uparrow$   & R@10$\uparrow$  & \multicolumn{1}{l}{MdR$\downarrow$} & \multicolumn{1}{l}{MnR$\downarrow$} \\ 
\midrule
DRL                               & ArXiv'22                                          &6.1	&30.6	&45.2	&13	&29.3  &4.5	&22.4	&38.0	&16	&26.4                   \\
Frozen                               & ICCV'21                                          &16.9	&56.4	&84.5	&5	&8.3  &18.9	&61.1	&83.1	&4	&8.8                   \\
CLIP4Clip                               & ArXiv'21                                          &28.6	&69.7	&89.7	&3	&6.2  &26.6	&72.0	&93.5	&3	&4.4                   \\
X-Pool                               & CVPR'22                                          & 27.2 & 68.6 & 87.2 & 3                   & 6.1                    & 32.8 & 70.9 & 93.6 & 3                   & 4.3                   \\
CenterClip                               & SIGIR'22                                          & 22.2 & 60.3 & 80.9 & 4                   & 9.1                    & 20.0 & 61.2 & 88.7 & 4                   & 7.1                   \\
X-Clip                               & ACMM'22                                          &29.0	&69.8	&88.6	&3	&6.4 &26.4	&74.4	&94.5	&3	&4.4                   \\
TS2-Net                               & ECCV'22                                          &29.3	&71.8	&90.9	&3	&5.8  &26.1	 &72.2	&95.1	&3	&4.4                   \\
UATVR                               & ICCV'23                                          &28.0	&68.5	&87.5	&3	&6.3  &25.1	&75.6	&94.4	&3	&4.1                   \\
UCoFiA                               & ICCV'23                                          &29.2 &71.3	&90.3	&3	&5.9  &28.1	&73.8	&94.3	&3	&4.2	                   \\
T-MASS                               & CVPR'24                                          &20.5	&55.5	&75.3	&5	&13.2  &23.6	&66.2	&85.8	&3	&6.1	                   \\
DGL                               & AAAI'24                                          & 26.5 & 65.4 & 85.1 & 3 & 7.1 & 27.2 & 70.1 & 91.5 & 3 & 5.2	                   \\
GLSCL                               & TIP'25                                          & 27.5 & 67.3 & 86.9 & 3 & 6.4 & 30.1 & 78.6 & 93.0 & 3 & 4.5	                   \\
TempMe                               & ICLR'25                                          & 28.7 & 70.6 & 89.5 & 3 & 6.0 & 32.5 & 80.1 & 94.0 & 2 & 4.3	                   \\
\rowcolor{gray!33}
Ours                               & $--$                                          &29.9	&71.2	&90.3	&3	&5.9  &33.8	&81.3	&94.8	&2	&4.2                   \\
\bottomrule
\bottomrule
\end{tabular}
\label{table 2}
\end{table*}

\begin{table*}[]
\centering
\caption{The experiments on the UMCRD drone video-text dataset compare MSAM with other state-of-the-art methods.}
\begin{tabular}{c|c|ccccc|ccccc}
\toprule
\toprule
\multirow{2}{*}{Methods} & \multirow{2}{*}{Date} & \multicolumn{5}{c|}{Text $\rightarrow$ Video}                                        & \multicolumn{5}{c}{Video $\rightarrow$ Text}                                       \\
                                & \multicolumn{1}{l|}{}                      & R@1$\uparrow$   & R@5$\uparrow$   & R@10$\uparrow$  & \multicolumn{1}{l}{MdR$\downarrow$} & \multicolumn{1}{l|}{MnR$\downarrow$} & R@1$\uparrow$   & R@5$\uparrow$   & R@10$\uparrow$  & \multicolumn{1}{l}{MdR$\downarrow$} & \multicolumn{1}{l}{MnR$\downarrow$} \\ 
\midrule
DRL                               & ArXiv'22                                          &9.3	&46.7	&64.0	&6	&25.0	 &10.6	&43.0	&61.4	&7	&24.1               \\
Frozen                               & ICCV'21                                          &28.4	&66.8	&80.2	&3	&11.6  &29.6	&64.3	&79.5	&3	&11.8                  \\
CLIP4Clip                               & ArXiv'21                                          &48.0	&82.9	&91.9	&2	&5.2	  &54.3	&90.0	&97.0	&1	&2.9	                   \\
X-Pool                               & CVPR'22                                          &42.8	&78.5	&89.3	&2	&6.2  &48.1	&86.3	&95.2	&2	&3.2	             \\
CenterClip                               & SIGIR'22                                          &38.2	&72.4	&84.4	&2	&9.0  &42.1	&80.9	&91.7	&2	&4.7                   \\
X-Clip                               & ACMM'22                                          &45.1	&81.3	&90.4	&2	&5.7  &51.4	&88.7	&95.1	&1	&3.0                   \\
TS2-Net                               & ECCV'22                                         &46.0	&82.0	&91.0	&2	&5.2  &53.4	&91.0	&95.7	&1	&2.8                   \\
UATVR                               & ICCV'23                                          &45.7	&80.5	&90.4	&2	&5.6	  &50.8	&88.2	&96.0	&1	&3.0	                   \\
UCoFiA                               & ICCV'23                                          &45.4	&81.8	&90.9	&2	&5.5  &53.2	&89.8	&96.4	&2	&2.8		            \\
T-MASS                               & CVPR'24                                          &39.9	&77.3	&88.7	&2	&6.2  &46.2	&85.4	&95.2	&2	&3.4               \\
DGL                               & AAAI'24                                          &41.0	&74.3	&85.7	&2	&7.6  &44.5	&81.2	&91.0	&2	&4.5	                   \\
GLSCL                               & TIP'25                                          & 44.2 & 78.6 & 88.1 & 2 & 6.4 & 49.0 & 86.2 & 93.3 & 2 & 3.5	                   \\
TempMe                               & ICLR'25                                          &46.5	&81.6	&90.8	&2	&5.3  &52.3	&89.0	&96.3	&1	&2.9	                   \\
\rowcolor{gray!33}
Ours                               & $--$                                          &49.5	&84.0	&92.2	&2	&4.8  &55.7	&91.7	&97.8	&1	&2.5                   \\
\bottomrule
\bottomrule
\end{tabular}
\label{table 3}
\end{table*}

\section{Experiments} \label{experiments}
This section details the datasets, evaluation metrics, and experimental setup, followed by a comparison of the proposed method with several advanced approaches on two datasets. Ablation studies and parameter tuning assess the impact of model components and structure. Finally, visualizations showcase the model's retrieval and semantic localization performance.

\subsection{Dataset and Evaluation Metrics}\label{Dataset and Evaluation Metrics}
In this study, two self-built drone video-text datasets are utilized, namely the UAV Scene Retrieval Dataset (USRD) and the UAV Multi-Context Retrieval Dataset (UMCRD). The UMCRD drone video-text dataset comprises 4,235 samples. 2,372 samples are collected on YouTube and 1,863 samples are collected by manual use of drones for filming. These samples encompass various urban surroundings, with images of varying sizes. On the other hand, the USRD drone video-text dataset consists of 2,864 samples from the ERA dataset \cite{DBLP:journals/corr/abs-2001-11394}, with each drone video sized at 640$\times$640 pixels.

Each drone video is paired with five text descriptions, as shown in the left column of Fig. \ref{fig:data}. The right column of Fig. \ref{fig:data} shows the word clouds of the USRD and UMCRD datasets. It can be observed that these datasets contain a variety of objects, with person, road, building, and plant being the most frequent object names. The most common color descriptors include green and red. In terms of spatial relationships, "middle" and "surround" are the most common descriptions. These descriptions are obtained through both manual annotation and automatic generation using the GPT large model.
\textbf{(1)} In the manual annotation process, different annotators provide information about the target’s color, size, and spatial relationships. Additionally, the annotators generate a title for the same video. Since each annotator may have a different understanding of the same video, this method helps produce more diverse titles. This not only enhances the inherent similarity between drone videos but also effectively increases the differentiation among similar videos.
\textbf{(2)} Inspired by ShareGPT4V \cite{DBLP:journals/corr/abs-2311-12793}, we use the Gemini API to generate descriptions for videos from the drone video dataset, creating a video-text pair dataset. We carefully design prompts to guide the Gemini model in generating detailed descriptions for drone videos, covering information such as the attributes, quantities, colors, shapes, and spatial positions of objects in the videos. These prompts ultimately result in structured, detailed, and comprehensive descriptions.

To analyze the datasets we constructed more thoroughly, Tab. \ref{table 1} shows a quantitative comparison between the two datasets. Although the USRD dataset contains 2,864 videos and the UMCRD dataset has 4,235 videos, this number is sufficient for the retrieval model to maintain robustness. We calculate the "Dataset Diversity" metric by comparing the number of completely distinct sentences in the dataset to the total number of videos. The scores for the USRD and UMCRD datasets are 3.43 and 4.86, respectively, indicating that UMCRD has greater title diversity. Additionally, we calculate the “Caption Ave Similarity" score, where a lower score signifies lower similarity between datasets, thus indicating higher diversity in descriptions. The scores for the USRD and UMCRD datasets are 0.026 and 0.032, respectively, suggesting that the text description diversity of USRD is slightly higher. We also tally the "Word Number" and "Caption Ave Length", finding that UMCRD scores higher than USRD in both metrics.

To maintain a fair assessment, the USRD dataset is partitioned into training, validation, and test sets, with 70\%, 20\%, and 10\% of the samples, respectively. For the UMCRD dataset, it is divided into training, validation, and test sets with 60\%, 20\%, and 20\% of the samples, respectively. By conducting experiments on these two datasets, the performance of our proposed method is comprehensively verified and analyzed in the DVTR task. We carry out two distinct video-text matching experiments: \textbf{(1)} "Video $\rightarrow$ Text" is the process of retrieving relevant textual information corresponding to a given query video. \textbf{(2)} "Text $\rightarrow$ Video" is the process of retrieving relevant videos corresponding to a given query text. 

We evaluate the performance of model using conventional retrieval metrics, including recall at rank K (R@K, higher is better), median rank (MdR, lower is better), and mean rank (MnR, lower is better). R@K refers to the proportion of test samples for which the correct result is found among the top K results retrieved. We present the results for R@1, R@5, and R@10. Median rank indicates the middle position of the correct result in the ranking, whereas mean rank represents the average position of all correct results.

\begin{table*}[]
\centering
\caption{The text-to-video results of the UMCRD and USRD drone video-text datasets are influenced by various factors.}
\begin{tabular}{c|ccccc|ccccc}
\toprule
\toprule
\multirow{2}{*}{Configurations} & \multicolumn{5}{c|}{UMCRD}   & \multicolumn{5}{c}{USRD}   \\ \cline{2-11} 

 & R@1$\uparrow$   & R@5$\uparrow$   & R@10$\uparrow$  & MdR$\downarrow$   & MnR$\downarrow$   & R@1$\uparrow$   & R@5$\uparrow$   & R@10$\uparrow$  & MdR$\downarrow$   & MnR$\downarrow$   \\ \midrule
\small{CLIP4Clip}                               &48.0	&82.9	&91.9	&2	&5.2  &28.6	&69.7	&89.7	&3	&6.2 \\
\small{+CIFFP}                               &49.1	&83.4	&92.0	&2	&5.0  &29.5	&70.1	&89.9	&3	&6.1 \\
++$\mathcal L_{ddsl}$                               &49.8	&84.1	&92.1	&2	&4.9  &29.7	&70.9	&90.3	&2	&5.9 \\
+++$\mathcal L_{dst}$                               &49.5	&84.0	&92.2	&2	&4.8  &29.9	&71.2	&90.3	&3	&5.9 \\
\bottomrule
\bottomrule
\end{tabular}
\label{table 4}
\end{table*}	

\subsection{Experiment Details}\label{Experiment details}
The MSAM method is implemented on a high-performance workstation using Pytorch and Mindspore frameworks. The workstation configuration includes a GeForce RTX 3090 GPU, an Intel Xeon(R) Silver 4210R CPU@2.40GHz with 40 cores, and 62.6 GiB of memory. To ensure fair comparison, all methods initialize both the image and text encoders with the pre-trained weights of CLIP's ViT-B/32. The projection dimensions of queries, keys, and values are set to $D$ = 512 to ensure consistency with the output dimensions of CLIP, and the logit scaling parameter from the CLIP pre-trained model is used for initialization. The hyper-parameter $\lambda$  is set to 0.1. The probabilistic embeddings $k$ is set to 7. All new projection weight matrices are initialized as identity matrices, and the bias terms are initialized to zero to build the model based on the text-image semantic understanding of the CLIP pre-trained model. The model undergoes end-to-end fine-tuning for each dataset. In all trials, the batch size is configured to 32, with a learning rate of 1e-6 applied to the CLIP-initialized weights, while the learning rate for other parameters is set to 1e-5. We use the AdamW optimizer \cite{DBLP:conf/iclr/LoshchilovH19} with a weight decay of 0.2 and apply cosine scheduling \cite{DBLP:conf/iclr/LoshchilovH17} for learning rate adjustment, following the approach of CLIP \cite{DBLP:journals/corr/abs-2104-08860}. In the experiments, we select 12 frames at regular intervals from each video and resize each frame to 224$\times$224, consistent with previous works \cite{DBLP:conf/iccv/BainNVZ21,DBLP:journals/corr/abs-2104-08860}.

\subsection{Method Comparison}\label{Method comparison}
To evaluate MSAM, we compare it with state-of-the-art methods, as detailed below:
\textbf{\sent Frozen\cite{DBLP:conf/iccv/BainNVZ21}}: Frozen efficiently leverages extensive image and video caption datasets to address challenges in visual architecture design and training noise.
\textbf{\sent DRL\cite{DBLP:journals/corr/abs-2203-07111}}: DRL enhances relevance with a Weighted Token Interaction module and improves hierarchical learning by reducing redundancy through Channel Decorrelation Regularization.
\textbf{\sent CLIP4Clip\cite{DBLP:journals/corr/abs-2104-08860}}: CLIP4Clip applies CLIP knowledge to video-text retrieval and verifies the role of image features.
\textbf{\sent X-Pool\cite{DBLP:conf/cvpr/GortiVMGVGY22}}: X-Pool integrates cross-modal attention to guide text in identifying semantically similar frames, enhancing video aggregation relevance.
\textbf{\sent CenterClip\cite{DBLP:conf/sigir/ZhaoZWY22}}: CenterClip segments videos and applies segment-level clustering to capture representative markers, lowering computation costs while enhancing semantic alignment.
\textbf{\sent X-Clip\cite{DBLP:conf/mm/MaXSYZJ22}}: X-Clip leverages attention over similarity matrices, emphasizing key frames and text, minimizing irrelevant interference, and improving cross-grained retrieval accuracy.
\textbf{\sent TS2-Net\cite{DBLP:conf/eccv/LiuXXCJ22}}: TS2-Net dynamically selects tokens across temporal and spatial dimensions to capture complete, fine-grained representations with subtle movement.
\textbf{\sent UATVR\cite{DBLP:conf/iccv/FangWLZS0SJW23}}: UATVR models text-video pairs as distributions in an optimized space, using adaptive tokens to capture multi-level semantics for complex queries.
\textbf{\sent UCoFiA\cite{DBLP:conf/iccv/WangSCBB23}}: UCoFiA captures scene and object-level cues, addressing multi-granularity cross-modal similarity for enhanced video retrieval.
\textbf{\sent T-MASS\cite{DBLP:conf/cvpr/WangWSLDRRT24}}: T-MASS employs similarity-aware radius and text regularization to improve text embedding flexibility and robustness beyond single-vector limitations.
\textbf{\sent DGL\cite{DBLP:conf/aaai/YangZWY24}}: DGL uses dynamic prompt tuning with global-local attention in a shared space, achieving strong video-text retrieval with few tunable parameters.
\textbf{\sent GLSCL\cite{DBLP:journals/tip/ZhangZGSDLS25}}: GLSCL aligns modalities via global interaction and shared queries, boosting retrieval through consistency and diversity losses.
\textbf{\sent TempMe\cite{DBLP:conf/iclr/ShenHHZZLBD25}}: TempMe reduces spatiotemporal redundancy via temporal token merging, boosting performance and speed with minimal parameters.

To assess the performance of the MSAM, we conducted systematic experimental comparisons on two constructed drone video-text datasets and calculated several evaluation metrics. As shown in Tab. \ref{table 2} and Tab. \ref{table 3}, the MSAM method outperforms all the comparison frameworks, fully demonstrating its advantage in leveraging richer visual contextual semantics. The experimental results demonstrate that our method outperforms current text-to-video retrieval methods on all datasets and evaluation metrics. Specifically, on the USRD dataset, our model improves the R@1 metric by 1.9\% compared to UATVR. Further analysis on the UMCRD dataset indicates that the retrieval task is more challenging due to the complexity of the text descriptions and the diversity of video content. Nevertheless, our method still surpasses UATVR with a 3.8\% improvement in R@1. This demonstrates the model's capability of dynamically aggregating semantic information, effectively matching the text with relevant video frames while suppressing irrelevant visual information from other video subregions. Additionally, on the UMCRD dataset, we successfully reduced the MdR metric to 2 and achieved the lowest MnR of 4.8, indicating that our model is more robust in handling erroneous retrieval samples. These findings further confirm the efficacy of our method in multi-semantic adaptive learning.

\subsection{Ablation Experiments}\label{Ablation experiments}
We perform a set of experiments to assess the impact of different influencing factors on the proposed MSAM method, as illustrated in Tab. \ref{table 4}. We use CLIP4Clip as the baseline method for experiments. Observations show that even without the multi-semantic adaptive learning mechanism, the results using CIFFP still outperform the baseline. This indicates that the interference from complex backgrounds in drone videos affects the learning performance of the model. Through the CIFFP method, we achieve joint reasoning between text and video frames, thereby enhancing the attention mechanism of the model. This method effectively filters out video frames that are unrelated to the text description. This allows the model to focus more on the core semantic content conveyed by the text, further optimizing the effectiveness of the supervision signal. Moreover, to address the issue of high intra-class similarity between different videos, MSAM introduces multiple semantic learning. It combines inter-frame variation information to extract specific regions from different frames, thereby constructing rich multi-semantic information that supports deep understanding and reasoning. Although there is a slight decrease in R@1 and R@5 in USRD after adding the $\mathcal L_{dst}$ constraint, the overall performance still improves. The $\mathcal L_{dst}$ constraint enhances the independence between features, reducing the instability caused by multicollinearity, thus maximizing the decoupling of distribution differences between features. In summary, all introduced learning modules have a positive impact on improving performance of the model.

\begin{figure}
\begin{center}
\includegraphics[width=0.49\textwidth]{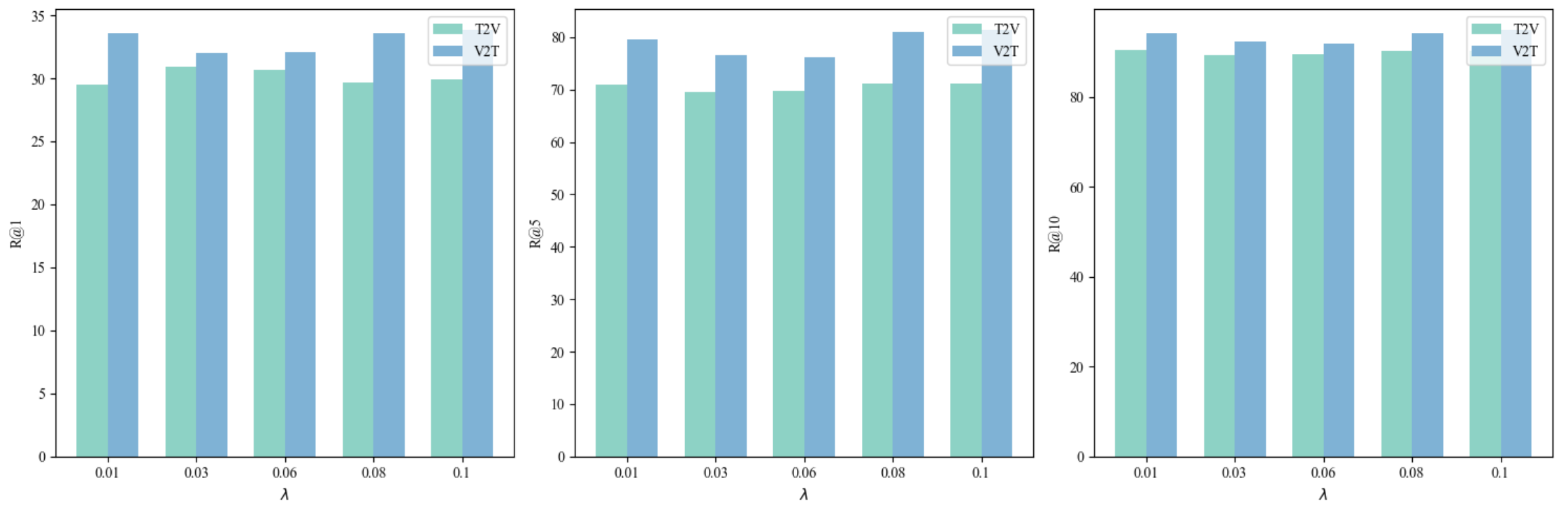}
\renewcommand{\figurename}{Fig.}
\end{center}
\caption{Visual analysis of the differences parameter $\lambda$ on USRD drone video-text dataset.}
\label{fig:param}
\end{figure}

\begin{table}[]
\centering
\caption{Performance of MSAM on USRD drone video-text dataset with differences between internal video and text embedding.}
\begin{tabular}{c|ccccc}
\toprule
\toprule
\multirow{2}{*}{Prob-Embs} & \multicolumn{5}{c}{Text $\rightarrow$ Video} \\ 
                  & R@1$\uparrow$   & R@5$\uparrow$   & R@10$\uparrow$  & MdR$\downarrow$ & MnR$\downarrow$   \\ \midrule
1                  &30.6	&69.9	&89.5	&3	&6.2   \\
3                  &30.7 &70.5 &88.9 &3 &6.2   \\
5                  &30.1 &70.5 &89.5 &3 &6.0   \\
7                  &29.9	&71.2	&90.3	&3	&5.9   \\
9                  &30.2 &69.3 &89.5 &3 &6.3   \\
\bottomrule
\bottomrule
\end{tabular}
\label{table 5}
\end{table}

\subsection{Parameter Analysis}\label{Parameter Analysis}

In Fig. \ref{fig:param}, we compare the recall rates under different values of $\lambda$. The results indicate that when $\lambda = 0.01$, the R@K score reaches its highest value, achieving a good balance between retrieval accuracy and recall. Although increasing $\lambda$ to 0.1 slightly decreases the R@K score, the overall performance improves. Further analysis reveals that a larger $\lambda$ enhances the model’s ability to learn multi-semantic representations through the loss term $\mathcal L_{dst}$, thereby improving its representational flexibility. In scenarios with low textual diversity, higher $\lambda$ values may yield better results. Overall, setting $\lambda = 0.1$ maintains high recall while further enhancing retrieval performance.

\begin{figure}
\begin{center}
\includegraphics[width=0.5\textwidth]{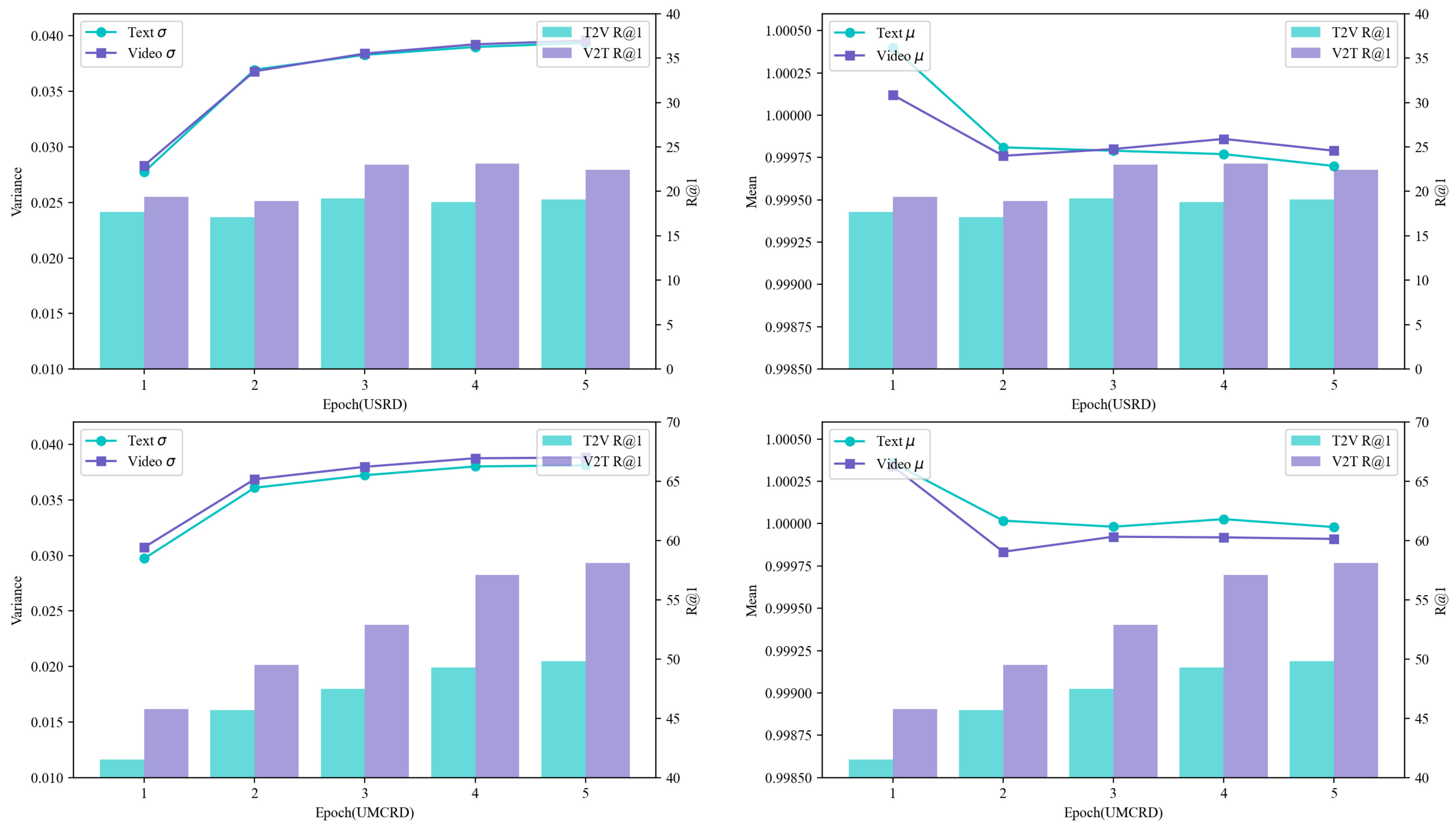}
\renewcommand{\figurename}{Fig.}
\end{center}
\caption{The differences between $k$ video and text embeddings generated by MSALM are visually analyzed using mean and variance. The legend illustrates their central tendency and variability.}
\label{fig:pn}
\end{figure}

\subsection{Analysis of Multi-Semantic Adaptive Learning Mechanism}\label{Analysis of Multi-Semantic Adaptive Learning Mechanism}
The MSALM increases the number of probabilistic embeddings. This enables more accurate modeling of the distribution of drone video and text features, although it also introduces additional computational overhead. In Tab. \ref{table 5}, we present the trend of text-to-video (t2v) retrieval performance as the number of sampled probabilistic embeddings ($k$) increases. The results show that as ($k$) increases, t2v retrieval performance gradually improves, but stabilizes when $k$=7. Considering computational cost, we ultimately choose $k$=7. Overall, the MSALM method significantly outperforms the baseline model, further validating the effectiveness of the MSALM. Fig. \ref{fig:pn} illustrates the impact of the MSALM on the variation trends of $k$ feature embeddings during the training process. By comparing two datasets, we studied the relationship between uncertainty and discriminability in the learned representations. Overall, as training progresses, the mean of the features gradually decreases. This may indicate that the model is learning more concentrated features, or that the values are approaching zero, suggesting that the embeddings are gradually converging. At the same time, the variance of the features increases, indicating that the distribution of embeddings within each video becomes more dispersed, with greater differences between frames or embeddings. We notice an enhancement in model performance, which further validates the efficacy of the MSALM.

\subsection{Analysis of Number of Frames Experiment}\label{Analysis of Number of Frames Experiment}
In our study, we initially sample 12 frames, aligning with current practices in text-video retrieval research \cite{DBLP:conf/cvpr/GortiVMGVGY22}. We extended our investigation by varying the frame sampling count in experiments conducted on the USRD and UMCRD datasets, across both training and inference phases, as detailed in Tab. \ref{table 6} and Tab. \ref{table 7}. The data indicates that increasing the frame count enhances model performance, presumably by enriching the information captured. However, once the frame count hits 12, we observe a stability in performance, with a minor downturn, potentially attributable to noise or extraneous details from an excess of frames. Importantly, the ideal frame sampling number is identified as a hyperparameter that is tailored to the specific dataset.

\begin{table}[]
\centering
\caption{Performance of MSAM on USRD drone video-text dataset with different visual frame numbers.}
\begin{tabular}{c|ccccc}
\toprule
\toprule
\multirow{2}{*}{Frames} & \multicolumn{5}{c}{Text $\rightarrow$ Video} \\ 
                  & R@1$\uparrow$   & R@5$\uparrow$   & R@10$\uparrow$  & MdR$\downarrow$ & MnR$\downarrow$   \\ \midrule
4                  &30.0	&69.7	&89.1	&3	&6.2    \\
8                  &30.5	&69.7	&89.3	&3	&6.1   \\
12                  &29.9	&71.2	&90.3	&3	&5.9   \\
16                  &29.6	&69.9	&89.1	&3	&6.2   \\
20                  &29.9	&69.6	&89.2	&3	&6.2   \\
\bottomrule
\bottomrule
\end{tabular}
\label{table 6}
\end{table}

\begin{table}[]
\centering
\caption{Performance of MSAM on UMCRD drone video-text dataset with different visual frame numbers.}
\begin{tabular}{c|ccccc}
\toprule
\toprule
\multirow{2}{*}{Frames} & \multicolumn{5}{c}{Text $\rightarrow$ Video} \\ 
                  & R@1$\uparrow$   & R@5$\uparrow$   & R@10$\uparrow$  & MdR$\downarrow$ & MnR$\downarrow$   \\ \midrule
4  &49.3	&83.7	&91.9	&2	&4.9  \\
8  &49.5	&83.9	&92.1	&2	&4.8   \\
12 &49.5	&84.0	&92.2	&2	&4.8     \\
16 &46.6	&82.9	&91.5	&2	&5.2  \\
20 &46.7	&82.5	&90.9	&2	&5.1  \\
\bottomrule
\bottomrule
\end{tabular}
\label{table 7}
\end{table}

\subsection{Analysis of Visualized Results}\label{Analysis of Visualized Results}
\begin{figure*}
\begin{center}
\includegraphics[width=0.9\textwidth]{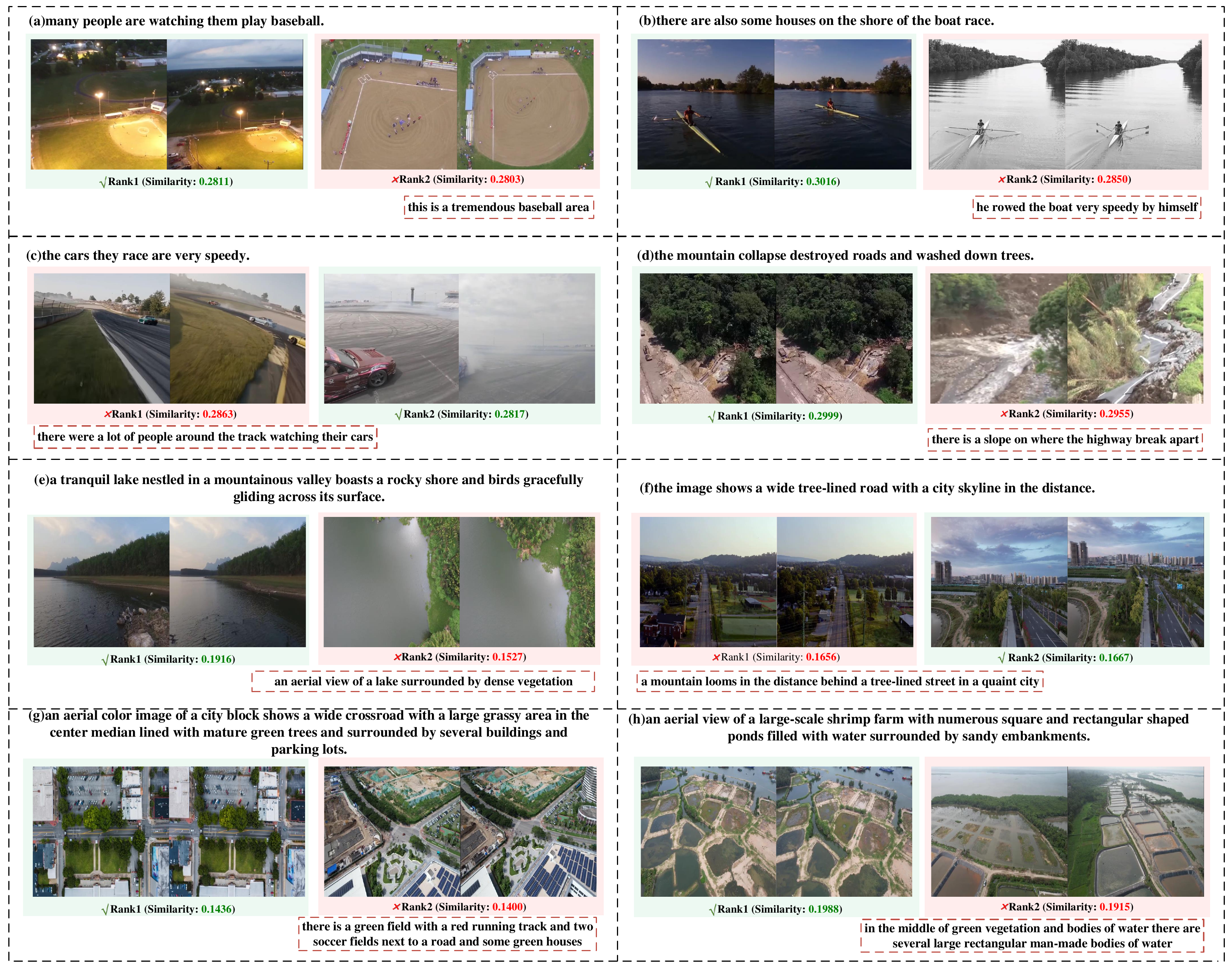}
\renewcommand{\figurename}{Fig.}
\end{center}
    \caption{The text-to-video retrieval results on the USRD (a-f) and UMCRD (e-h) datasets are presented. The query descriptions for each group are located in the top left corner. Each pair of frames showcases content extracted from the target video. Correct results are marked in green, incorrect ones in red, with correct descriptions of wrong results shown below.}
\label{fig:results}
\end{figure*}

In this section, we present the visualization of the retrieval outcomes. Fig. \ref{fig:results} shows the top two text-video retrieval results obtained by inputting text queries. As shown in Fig. \ref{fig:results}(a), the query results exhibit high similarity, indicating that our proposed method effectively retrieves segments from the video library that match the query content. For example, for the query "many people are watching them play baseball", our model accurately identifies the baseball game scene and matches it with the query, showing a high similarity score. However, we also face some challenges. As shown in Fig. \ref{fig:results}(c), for the query "the cars they race are very speedy", the ambiguity in the definition of "speedy" leads to the ground truth being ranked second, although the first result identified by the model better fits the query. The model determines the speed by comparing the motion changes of different vehicles. Additionally, the model performs well in handling fine details. For instance, in Fig. \ref{fig:results}(e) with the query "small bird" and in Fig. \ref{fig:results}(h) with the query "sandy embankment", the model successfully captures these details. This is attributed to the multi-semantic adaptive learning mechanism, which enables the model to learn more fine-grained distinctions. As shown in Fig. \ref{fig:results}(d) and Fig. \ref{fig:results}(f), some query results have lower similarity scores, likely due to subtle differences between video contents that make discrimination more difficult. Although these retrieved videos include landslides and traffic road scenes, the high similarity between video frames leads to less ideal similarity scores. The experimental results indicate that multi-semantic adaptive learning enhances ability of the model to adapt to different data scenarios and queries. CIFFP allows the visual model to better understand textual cues, thereby improving localization accuracy.

\subsection{Analysis of Semantic Location Visualization}\label{Analysis of Semantic location visualization}
\begin{figure*}
\begin{center}
\includegraphics[width=0.9\textwidth]{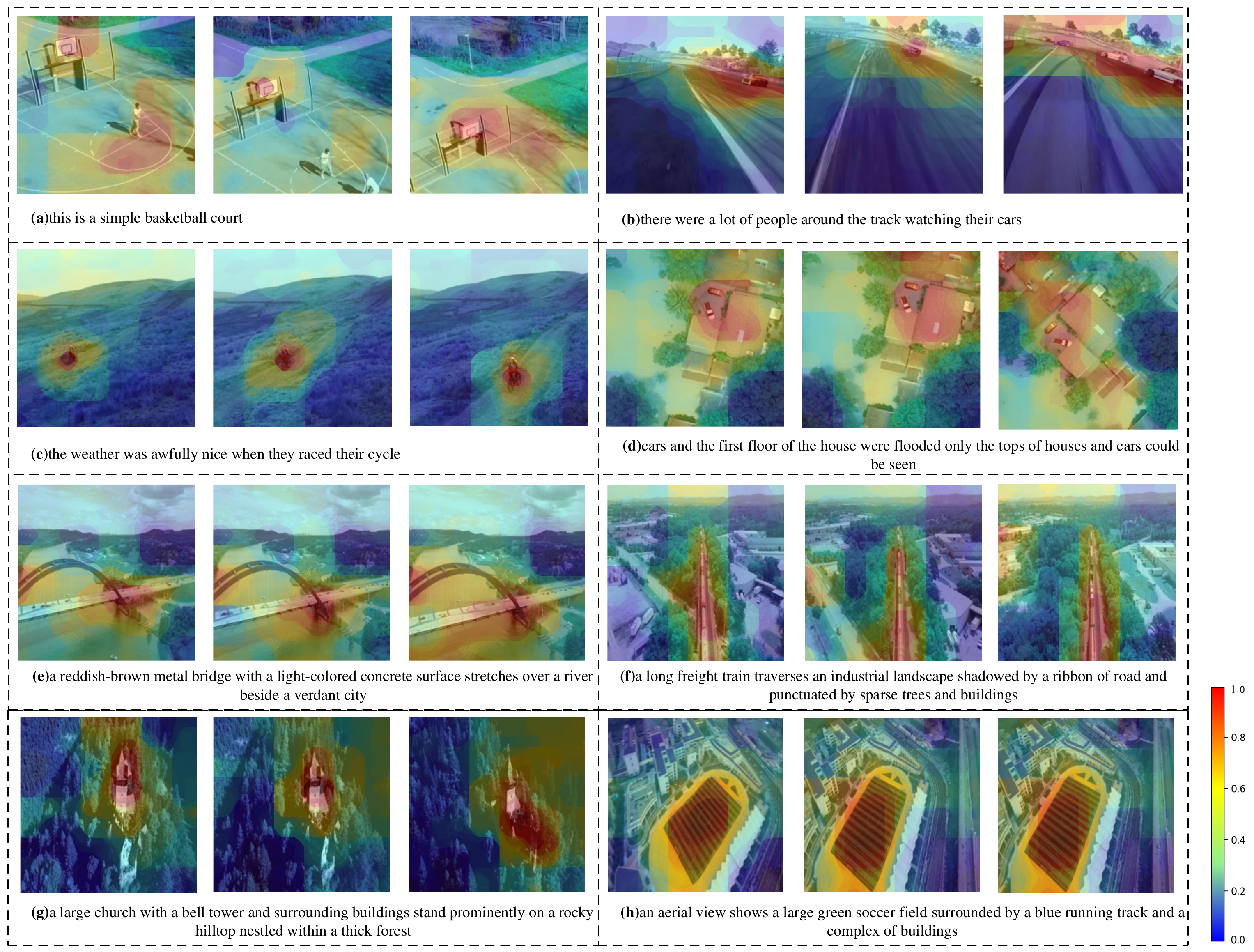}
\renewcommand{\figurename}{Fig.}
\end{center}
\caption{The visualization displays the heatmap relationship between the query text and the intermediate frames of the corresponding drone video. Panels (a) to (h) present the visual results for eight representative drone videos. Darker red indicates higher attention, while darker blue indicates lower attention.}
\label{fig:dr}
\end{figure*}

In this study, we select three frames from each video for each example and analyze in detail how the model distributes its attention from the given text to each frame. We also introduce saliency region masks to highlight the key areas that the model focuses on. Fig. \ref{fig:dr} shows the saliency mask results of eight typical videos. In Fig. \ref{fig:dr}(a), the video is described as "this is a simple basketball court". We notice that the model mainly concentrates on the key regions associated with the basketball court. Additionally, the model also clearly pays attention to details such as the people present. In Fig. \ref{fig:dr}(b), the text description is "there were a lot of people around the track watching their cars". The model specifically emphasizes the crowd and vehicles on the track, indicating that the algorithm tends to infer that people are more likely to describe the relationship between the people and the cars. In Fig. \ref{fig:dr}(e), the text query is "a reddish-brown metal bridge with a light-colored concrete surface stretches over a river beside a verdant city". The results indicate that the model predominantly focuses on the river and the bridge, with particular emphasis on the region surrounding the bridge. This example shows that the model infers that people are more likely to describe the relationship between the bridge and the river. Nevertheless, in contrast to the other two scenes, the model appears to focus less on the buildings in the video, likely perceiving the river and bridge as more visually prominent. In Fig. \ref{fig:dr}(f), the query phrase is "a long freight train traverses an industrial landscape shadowed by a ribbon of road and punctuated by sparse trees and buildings". The model accurately highlights the freight train on the road, aligning with the description. However, the attention to the industrial area is relatively limited. Based on the visualized outcomes, we observe that the model excels in the reasoning task, successfully emphasizing important objects without the need for extra information. While there is still room for improvement compared to existing visual localization techniques, this method shows promising potential for further development with more in-depth research.

\subsection{Analysis of Video Frame Pooling Methods}\label{Analysis of Video Frame Pooling Methods}
In this section, we evaluate our CIFFP pooling method against mean pooling \cite{DBLP:journals/corr/abs-2104-08860}, Top-k pooling \cite{DBLP:conf/cvpr/GortiVMGVGY22}, Self-Attention pooling, and X-Pool \cite{DBLP:conf/cvpr/GortiVMGVGY22} for text-to-video retrieval, as presented in Tab. \ref{table 8}. The results demonstrate that our method achieves the best overall performance. The main reason for this is that mean pooling does not incorporate text information, making it ineffective in guiding the learning process of the model. While both Top-k pooling, Self-Attention pooling and X-Pool pooling methods consider text information, they rely too heavily on the text, overlooking important information learned from the visual modality. This issue becomes more pronounced as the number of scenes in drone videos increases, leading to the aggregation of irrelevant elements. Conversely, our CIFFP effectively extracts the visual information related to the text descriptions from drone video frames by integrating textual information and enhancing the inter-frame visual relationships, improving retrieval performance.

\begin{table}[]
\centering
\caption{Performance of MSAM on USRD drone video-text dataset with different pooling methods.}
\begin{tabular}{c|ccccc}
\toprule
\toprule
\multirow{2}{*}{Pooling} & \multicolumn{5}{c}{Text $\rightarrow$ Video} \\ 
                  & R@1$\uparrow$   & R@5$\uparrow$   & R@10$\uparrow$  & MdR$\downarrow$ & MnR$\downarrow$   \\ \midrule
Mean  &28.1	&68.8	&88.3	&3	&6.1  \\
Top-k  &27.2	&68.5	&88.3	&3	&6.2   \\
Self-Atten  &25.7	&64.7	&84.9	&3	&6.8   \\
X-Pool & 27.2 & 68.6 & 87.2 & 3                   & 6.1     \\
CIFFP &27.3	&69.1	&88.4	&3	&6.0  \\
\bottomrule
\bottomrule
\end{tabular}
\label{table 8}
\end{table}

\subsection{Analysis of Time Consumption and Parameter Quantity}\label{Analysis of Time Consumption and Parameter Quantity}
Tab. \ref{table 9} shows the time complexity and parameter scale of each module, with the feature dimension $D$ omitted for simplicity. Although the computational complexity of DDSL and DS is slightly higher, the overall cost per drone video-text pair remains within an acceptable quadratic range. Notably, the CIFFP module improves matching performance without increasing model complexity. Overall, the MSAM method maintains a simple structure while demonstrating significant performance improvements.

\begin{table}[]
\centering
\caption{Results of parameter quantity and evaluation time for different components. $N$ denotes sample size. $k$ is the number of probabilistic embeddings.}
\begin{tabular}{c|c|c}
\toprule
\toprule
Methods & Time Complexity & Params(M) \\ \midrule
CLIP4Clip & $\mathcal{O}(N^{2})$ & 162.3 \\
w/ CIFFP & $\mathcal{O}(N^{2})$ & 164.4 \\
w/ ASC & $\mathcal{O}(N^{2})$ & 165.3 \\
w/ DDSL & $\mathcal{O}(N^{2}k^{2})$ & 167.3 \\
w/ DS & $\mathcal{O}(N^{2}k^{2})$ & 167.3 \\
\bottomrule
\bottomrule
\end{tabular}
\label{table 9}
\end{table}

\section{Conclusions}\label{Conclusion}
In this paper, we construct two new cross-modal drone video-text datasets and conduct an in-depth exploration of the complex semantic features within drone videos. To achieve this, we propose a method called MSAM, based on a multi-semantic adaptive learning mechanism. It combines dynamic changes between frames to extract rich semantic information from specific scene areas, enhancing deep understanding and reasoning of drone video content. The cross-modal interactive feature fusion pooling mechanism uses textual cues to select visual features that correspond to the text descriptions. This enables the exploration of detailed connections relationships between drone video frames and texts. Additionally, we introduce multiple objective functions to provide stronger guidance for visual semantics and enhance the interactions between modalities. All these modules work together to align the drone video and text into a unified embedding space. Comprehensive experimental findings highlight the considerable effectiveness of this framework on two cross-modal drone video-text retrieval datasets.

\ifCLASSOPTIONcaptionsoff
  \newpage
\fi

\bibliographystyle{IEEEtran}
\bibliography{Refer}

\end{document}